\documentclass[10pt,journal,compsoc]{IEEEtran}

\ifCLASSOPTIONcompsoc
  \usepackage[nocompress]{cite}
\else
  \usepackage{cite}
\fi
\ifCLASSINFOpdf
\else
\fi
\usepackage{cite}
\usepackage{multirow}
\usepackage{graphicx}
\usepackage{subfigure}
\usepackage{color}
\usepackage{dcolumn}
\usepackage{amsmath}
\usepackage{algorithm}
\usepackage{algorithmic}
\usepackage{verbatim}
\usepackage{float}
\usepackage{comment}

\begin{document}

\title{Hardware-Guided Symbiotic Training for Compact, Accurate, yet Execution-Efficient LSTM}

\author{Hongxu Yin, Guoyang Chen, Yingmin Li, Shuai Che, Weifeng Zhang, and Niraj K. Jha,~\IEEEmembership{Fellow,~IEEE}
\thanks{Hongxu Yin and Niraj K. Jha are with the Department
of Electrical Engineering, Princeton University, Princeton,
NJ, 08544 USA, e-mail:\{hongxuy, jha\}@princeton.edu. Guoyang Chen, Yingmin Li, Shuai Che, and Weifeng Zhang are with Alibaba Group US Inc., Sunnyvale, CA, 94085 USA, e-mail:\{g.chen, yingmin.li, shuai.che, weifeng.z\}@alibaba-inc.com.}}

\IEEEtitleabstractindextext{%
\begin{abstract}
Many long short-term memory (LSTM) applications need fast yet compact models. 
Neural network compression approaches, such as the grow-and-prune paradigm, 
have proved to be promising for cutting down network complexity by skipping 
insignificant weights. However, current compression strategies are mostly 
hardware-agnostic and network complexity reduction does not always translate 
into execution efficiency. In this work, we propose a hardware-guided symbiotic 
training methodology for compact, accurate, yet execution-efficient inference 
models. It is based on our observation that hardware may introduce substantial 
non-monotonic behavior, which we call the latency hysteresis effect, when 
evaluating network size vs. inference latency. This observation raises question
about the mainstream smaller-dimension-is-better compression strategy, which 
often leads to a sub-optimal model architecture. By leveraging the 
hardware-impacted hysteresis effect and sparsity, we are able to achieve the 
symbiosis of model compactness and accuracy with execution efficiency, thus 
reducing LSTM latency while increasing its accuracy. We have evaluated our 
algorithms on language modeling and speech recognition applications. Relative 
to the traditional stacked LSTM architecture obtained for the Penn Treebank 
dataset, we reduce the number of parameters by 18.0$\times$ (30.5$\times$) 
and measured run-time latency by up to 2.4$\times$ (5.2$\times$) on Nvidia 
GPUs (Intel Xeon CPUs) without any accuracy degradation.  For the DeepSpeech2 
architecture obtained for the AN4 dataset, we reduce the number of parameters 
by 7.0$\times$ (19.4$\times$), word error rate from 12.9\% to 9.9\% (10.4\%), 
and measured run-time latency by up to 1.7$\times$ (2.4$\times$) on Nvidia GPUs (Intel Xeon CPUs). Thus, our method yields compact, accurate, yet 
execution-efficient inference models.
\end{abstract}

}

\maketitle

\IEEEdisplaynontitleabstractindextext

\IEEEpeerreviewmaketitle

\renewcommand{\thefootnote}{}

\IEEEraisesectionheading{\section{Introduction}}

Long short-term memory (LSTM) has been widely deployed for applications 
like speech recognition~\cite{deepspeech2}, neural machine 
translation~\cite{seq2seq}, health monitoring~\cite{deepheart}, and language 
modeling~\cite{stanford,wenwei}. It is capable of learning both the long-term 
and short-term dependencies in sequential data~\cite{lstm}. Researchers have 
kept increasing the depth and size of LSTM models to improve their accuracy. 
For example, the DeepSpeech2 architecture \cite{deepspeech2} is more than 
2$\times$ deeper and 10$\times$ larger than the initial DeepSpeech architecture
\cite{deepspeech1}. A deep neural network (NN) architecture allows the model to 
capture low/mid/high-level features through a multi-level abstraction that 
typically results in high inference accuracy~\cite{him}. 
But it also leads to a sharp increase in computation, thus posing significant 
challenges to model deployment. In addition, a large NN model 
consumes substantial storage, memory bandwidth, and computational resources 
that may be too excessive for mobile and embedded 
devices~\cite{ternary, shift, iot_energy, multicoset}. Furthermore, the 
increasingly stringent latency constraints imposed by real-time applications 
make large high-latency LSTMs unusable. Thus, it is practically important to 
optimize the model from all three aspects of performance simultaneously: 
model compactness, accuracy, and execution efficiency.

Network compression has emerged as a promising technique to reduce the 
computational cost of deep NNs by eliminating connections
with insignificant weights, such as zeros or near-zeros. By leveraging 
effective network growth~\cite{nest} and pruning~\cite{PruningHS} techniques, 
the number of parameters can be cut down by over 30$\times$ for convolutional 
neural networks (CNNs)~\cite{PruningHS,nest,admm} and more than 10$\times$ for 
LSTMs~\cite{ese,wenwei,baiduprune,hlstm}. However, current compression 
strategies are mostly hardware-agnostic, and network complexity reduction does 
not always translate into execution efficiency and may even have an adverse
impact on other performance metrics. For example, training NNs 
towards extreme weight sparsity offers little execution performance gain on 
current GPUs due to a lack of effective sparsity support. Moreover, some 
compressed networks may even suffer from inefficient execution, as observed 
in~\cite{scalpel}.

In this work, we propose a novel hardware-guided symbiotic training methodology 
based on our observation that the hardware may introduce substantial 
non-monotonicity (we call this the {\em latency hysteresis effect (LHE)}): 
a smaller model, which typically has a lower accuracy, may also be 
slower at run-time. This observation 
raises a question about the mainstream smaller-dimension-is-better
strategy, which often leads to a sub-optimal design point in the model 
architecture space. By leveraging the hardware-impacted hysteresis effect, we 
are able to achieve the symbiosis of all three performance aspects: higher 
accuracy, smaller model size, and lower inference latency. To evaluate this 
symbiotic strategy, we adopt the internally deeper hidden-layer LSTM (H-LSTM) 
structure \cite{hlstm} to reduce the number of stacked layers, and start 
training from a sparse seed architecture, which grows effective connections to 
reach an initial high accuracy. Then we employ our hardware-guided structured 
grow-and-prune algorithms to shrink the network into hardware-favored 
dimensions. Finally, we prune the network weights again for extra compactness. 

The major contributions of our approach can be summarized as follows:
\begin{enumerate}
  \item We propose a novel training methodology to exploit hardware 
LHE to achieve a symbiosis of model compactness and accuracy with
reduced run-time latency.
  \item We combine multi-granular grow-and-prune algorithms with hardware 
guidance to reduce the model into a hardware-favored architecture. This is 
the first work that effectively avoids sub-optimal design points that may 
consume as much as 90\% of the model architecture space.
  \item The reported results outperform those from the literature from all 
three design perspectives: (a) 7.0$\times$ to 30.5$\times$ model compression, 
(b) higher accuracy, and (c) 1.4$\times$ to 5.2$\times$ reduction in run-time 
latency on Nvidia GPUs and Intel Xeon CPUs. Thus, our method yields compact, 
accurate, yet execution-efficient inference models.
\end{enumerate}

The rest of this paper is organized as follows. We review related works in 
Section 2. Then, we explain the motivation of this work in Section 3. In 
Section 4, we discuss our proposed methodology in detail. We present our 
experimental results on both language modeling and speed recognition in 
Section 5. Finally, we draw a conclusion in Section 6.

\section{Related work}
Various attempts have been made to improve the efficiency of LSTM models. 
One direction focuses on improving the LSTM cells. The gated recurrent unit 
(GRU) utilizes reset and update gates to achieve a similar performance to
an LSTM while reducing computational cost~\cite{gru}. Quasi-RNN explores the 
intrinsic parallelism of time series data to outperform an LSTM for the same 
hidden state width~\cite{qrnn}. H-LSTM incorporates deeper control gates to 
reduce the number of external stacked layers. It achieves higher accuracy 
than the GRU and LSTM with fewer parameters~\cite{hlstm}. 

Network compression techniques, such as the grow-and-prune paradigm, have 
recently emerged as another direction for reducing LSTM redundancy. The 
pruning method was initially shown to be effective on large CNNs by
demonstrating the reduction in the number of parameters in AlexNet by 9$\times$ 
and VGG by 13$\times$ for the well-known ImageNet dataset, without any accuracy 
loss~\cite{PruningHS}. Follow-up works have successfully scaled this technique 
to LSTMs~\cite{ese,wenwei,baiduprune}. For example, a recent work proposes 
structured pruning for LSTMs through group LASSO regularization~\cite{wenwei}. 
Network growth is a complementary method to pruning.
It enables a more sparse yet accurate model to be obtained before pruning 
starts~\cite{nest}. A grow-and-prune paradigm typically reduces the 
number of parameters in CNNs \cite{nest} and LSTMs \cite{hlstm} by another 
2$\times$.  However, all these methods are hardware-agnostic. Most of them 
utilize monotonic optimization metrics, e.g., smaller matrix dimensions or 
fewer multiply-accumulate operations, hence optimize towards slimmer or more 
sparse models that may not necessarily translate into execution efficiency.

\begin{figure}[t]
\begin{center}
\includegraphics[width=7.9cm]{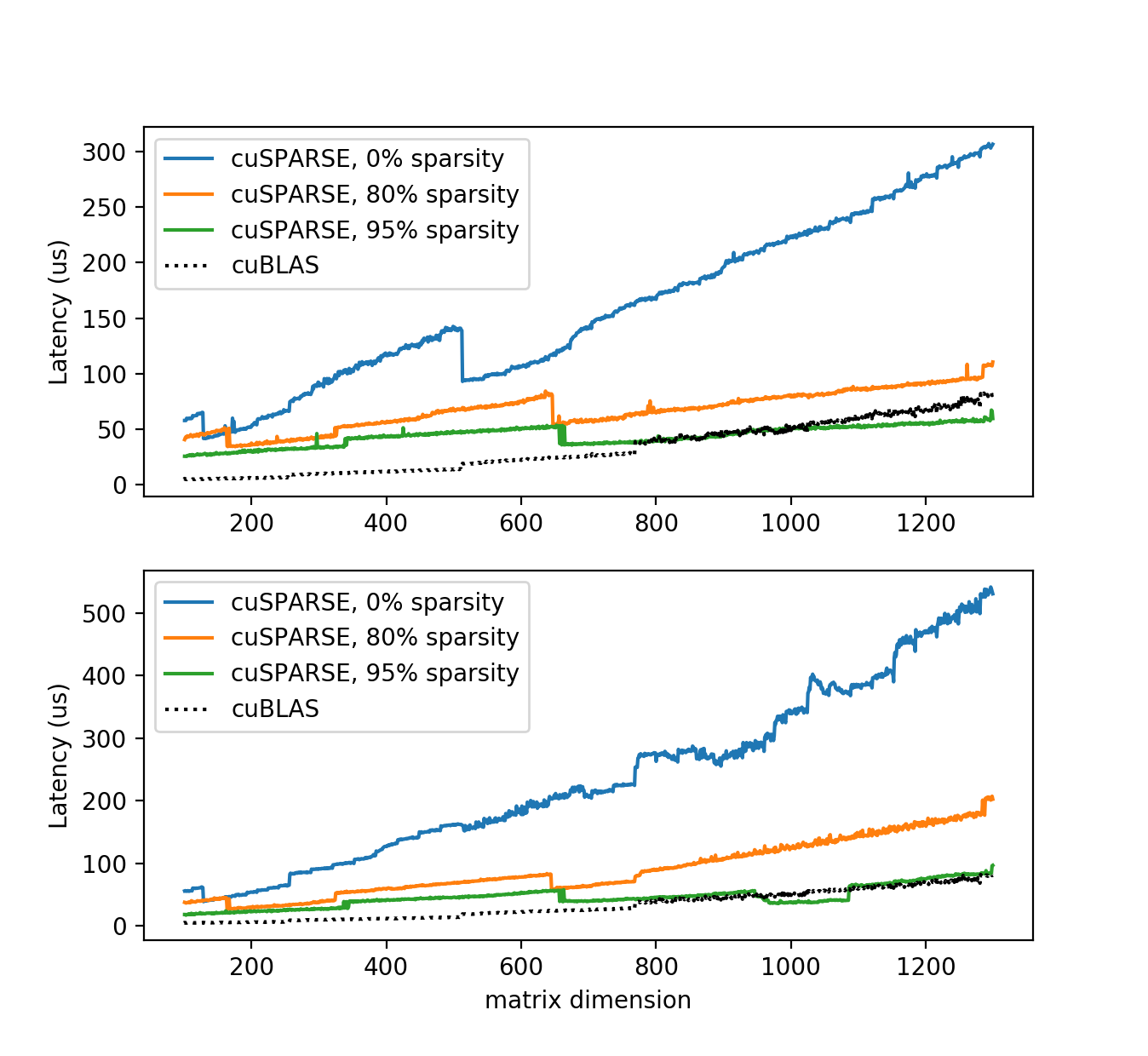}
\end{center}
\caption{Matrix multiplication latency on Nvidia Tesla P100 (up) and Quadro 
P2000 (down) GPUs using cuSPARSE and cuBLAS libraries. Weight matrix: sparse
and square, dimension on the $x$-axis. Input matrix: dense, batch size 16.}
\vspace{-5mm}
\label{fig:cusparse_vs_cublas}
\end{figure}

There have been some recent efforts towards bridging the gap between complexity 
removal and execution efficiency for CNNs through hardware-heuristics-guided 
pruning approaches. For example, Scalpel~\cite{scalpel} adopts different 
pruning strategies based on three hardware parallelism levels (low, moderate, 
and high) of the underlying hardware.  DeftNN~\cite{deftnn} removes a synapse 
vector that is highly correlated with another one in the weight matrix, on the 
assumption that a smaller dimension leads to improved latency. However, both 
works are based on high-level hardware heuristics rather than real hardware 
behavior like LHE, which may lead to sub-optimal networks. Energy-aware 
pruning~\cite{mitenergy} adopts a hardware energy consumption model in its 
pruning criteria. However, it leads to a 
re-design for each target hardware and requires expert knowledge of the 
hardware. Thus, it is not very user-friendly. Chameleon~\cite{chamnet}
can effectively adapt CNNs to target platforms and deliver ChamNets that
achieve consistent accuracy gains across various latency constraints
relative to MobileNetV2~\cite{mobilenetv2}, MnasNet~\cite{mnasnet}, and 
ShuffleNetV2~\cite{shufflenetv2}. However, construction of Chameleon's three 
predictors may require training hundreds of baseline models, hence may be 
time-consuming.  In the domain of recurrent NNs, a relevant work 
explores hardware-inspired weight- and block-level sparsity for speech 
recognition~\cite{narangblock}. 
With cuSPARSE library support, it delivers 0.8$\times$ to 4.5$\times$ speed-up 
relative to the dense baseline. However, we find cuSPARSE~\cite{cusparse} to 
be slower than the latest cuBLAS~\cite{cublas} dense 
library for matrix multiplication, which is a key 
operation in LSTM. In Fig.~\ref{fig:cusparse_vs_cublas}, we compare these two 
libraries on GPUs over a typical dimension range of LSTMs. It can be observed 
that cuSPARSE is slower than cuBLAS even at a 95\% sparsity level.

\section{Motivation}

As opposed to most prior works that use floating-point operations
(FLOPs) or multiply-and-accumulate (MAC) operations as an indirect metric 
for evaluating model compactness, we aim to develop an automated LSTM 
synthesis flow that acts on directly measured inference latency. This
flow does not adopt the traditional assumption that a smaller model 
(e.g., with smaller hidden state widths) is implicitly faster. In fact,
we show that such an assumption is often not valid at run-time on
hardware. This points to the need for a new methodology that can link model 
simplification algorithms to direct execution benefits.

\begin{figure}[t]
\begin{center}
\includegraphics[width=\columnwidth]{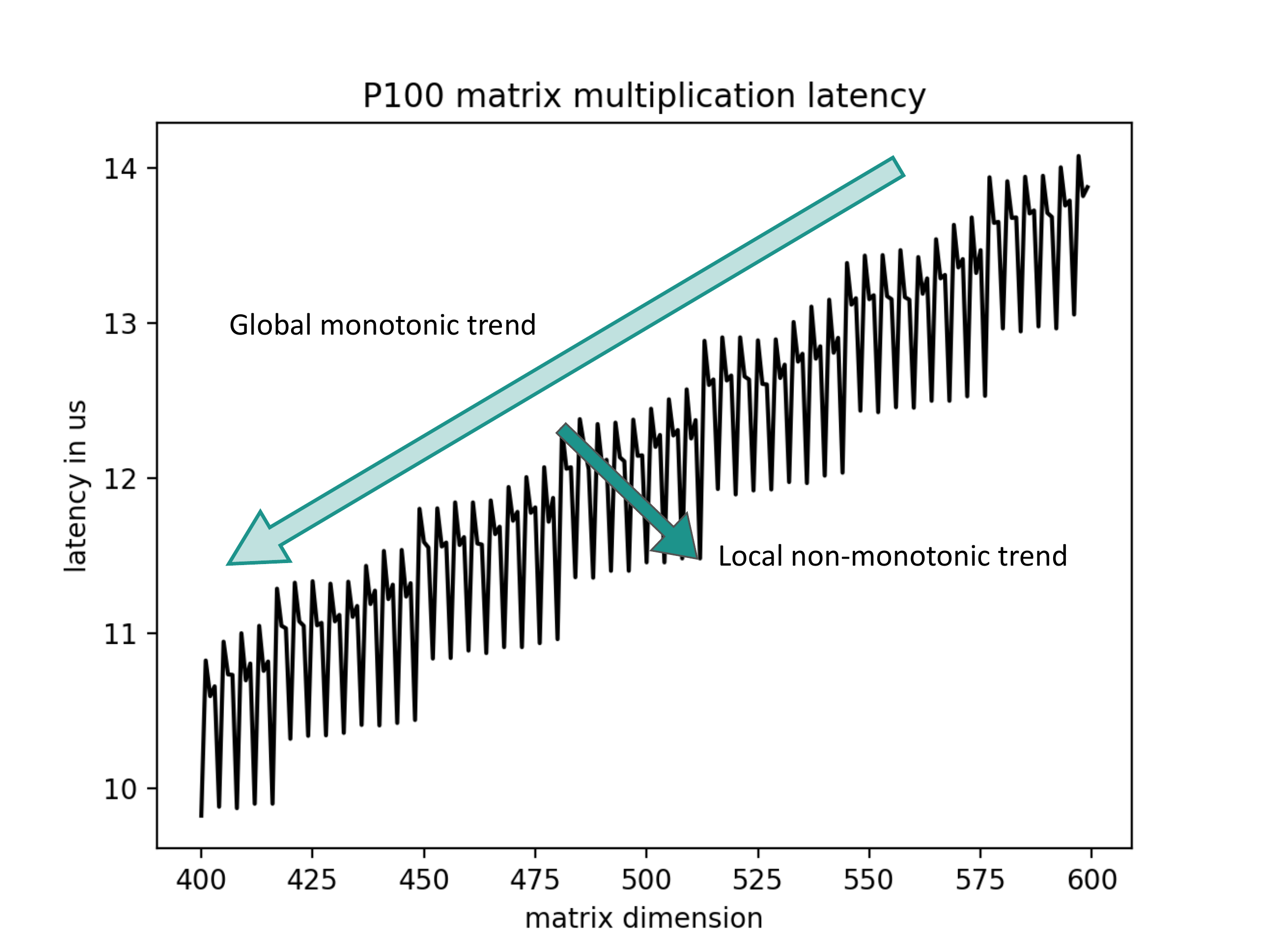}
\end{center}
\caption{Example of non-monotonicity in matrix multiplication on the Nvidia 
Tesla P100 GPU. The weight matrix is square with its dimension on the
$x$-axis. The input has a batch size of 16. } 
\label{fig:mm_nonlinear}
\end{figure}

Let us first profile the latency of the matrix multiplication operation on 
a GPU, as shown in Fig.~\ref{fig:mm_nonlinear}, due to its computational 
importance. This operation consumes more than half of the computational time 
in LSTMs. We observed two distinct trends when considering matrix dimension 
vs.~latency:

\begin{enumerate}
\item Global monotonic trend: a smaller dimension is, in general, faster in 
terms of run-time latency due to the reduced number of weights 
(i.e., computation).
\item Local non-monotonic trend: the run-time latency lags behind or even 
reverts the trend as the weight dimension decreases. We refer to this local 
trend as LHE and the point where LHE starts to occur as the latency 
hysteresis point (LHP). Within the latency hysteresis bin (i.e., the local 
range), smaller dimensions worsen run-time latency relative to the 
corresponding LHP.
\end{enumerate}

\begin{figure}[h]
\begin{center}
\includegraphics[width=\columnwidth]{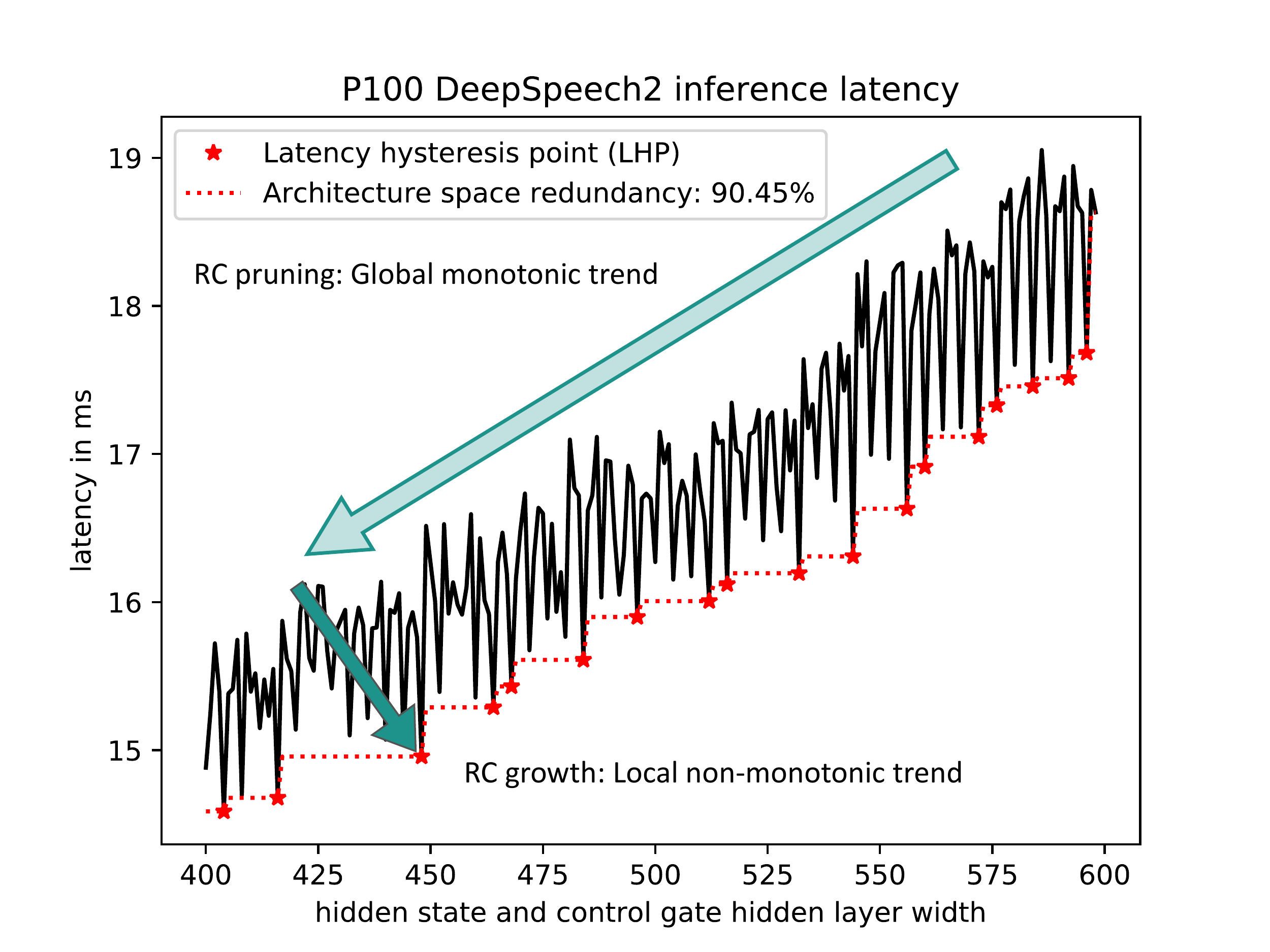}
\end{center}
\caption{Inference latency per test instance for the DeepSpeech2 architecture 
on the Nvidia Tesla P100 GPU. The hidden state and control gate hidden layer 
widths are set to the same number. The input has a batch size of 16. 
Architecture space redundancy: ($1-$ \#LHPs / \#total design points)$\times100\%$. RC: row/column.}
\label{fig:model_nonlinear}
\end{figure}

LHE is caused by cache line granularity when loading/storing data and 
vectorization optimization (e.g., vectorized vs. general matrix multiplication 
kernels) enabled at some particular data input dimensions to take full 
advantage of the bus bandwidth and single-instruction-multiple-data (SIMD) 
nature of hardware processing units. Change of optimization strategies for 
memory placement or computation scheduling can easily impact the final 
execution efficiency for inference~\cite{porple, effisha}. 
This also leads to a pervasive presence of LHE on a CPU.

The impact of LHE can scale up to the inference model level. For example, 
Fig.~\ref{fig:model_nonlinear} shows a plot of the DeepSpeech2 inference 
latency against model size (specified by the hidden state width and the 
control gate hidden layer width). We can observe that a smaller DeepSpeech2 
architecture, which also typically has a lower accuracy, may also be 
slower in run-time. This raises a question about the mainstream 
smaller-is-better strategy, given the existence of a large number of LHPs that 
make more than 90\% of the design points in Fig.~\ref{fig:model_nonlinear} sub-optimal. 

\section{Methodology}

\begin{figure}[t]
\begin{center}
\includegraphics[width=7.0cm]{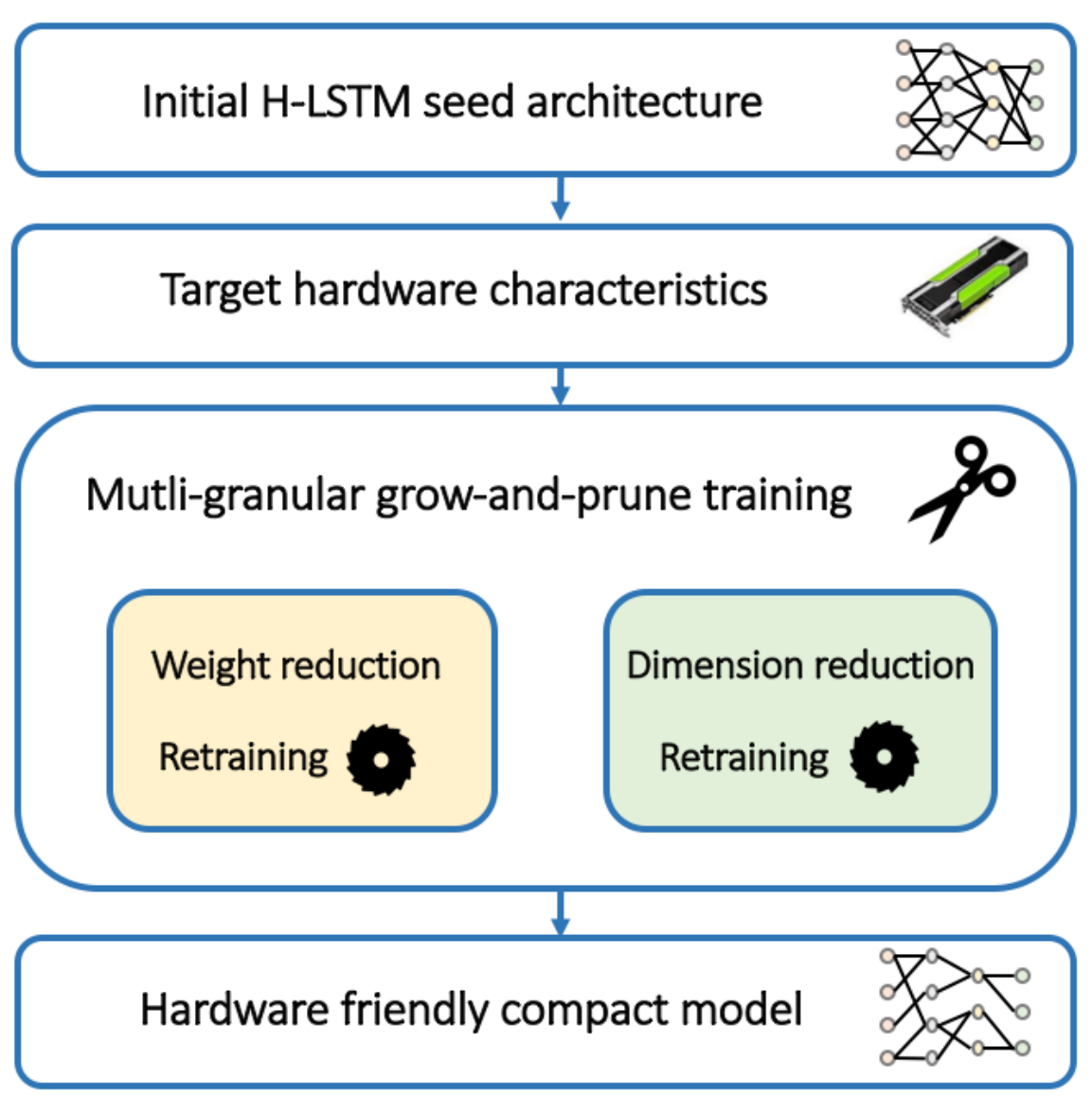}
\end{center}
\caption{An illustration of the hardware-aware architecture synthesis flow.}
\label{fig:flow_chart}
\end{figure}

In this work, we combine the multi-granular grow-and-prune algorithms with 
network profiling to optimize the network towards a joint algorithm-hardware 
optimal solution. We summarize our main synthesis flow in 
Fig.~\ref{fig:flow_chart}.  The proposed synthesis process starts with a 
partially connected seed architecture. Under the guidance of hardware 
profiles, the multi-granularity of our algorithms enables the network to 
adaptively expand (row/column growth), shrink (row/column pruning), 
condense (weight growth), and sparsify (weight pruning) into hardware-favored 
design points. This benefits hardware even when there is no sparsity support. 
In the final stage, the synthesis flow rests at a compact design point that 
is both compact and hardware friendly.

\begin{figure*}[h]
\begin{center}
\includegraphics[width=\linewidth]{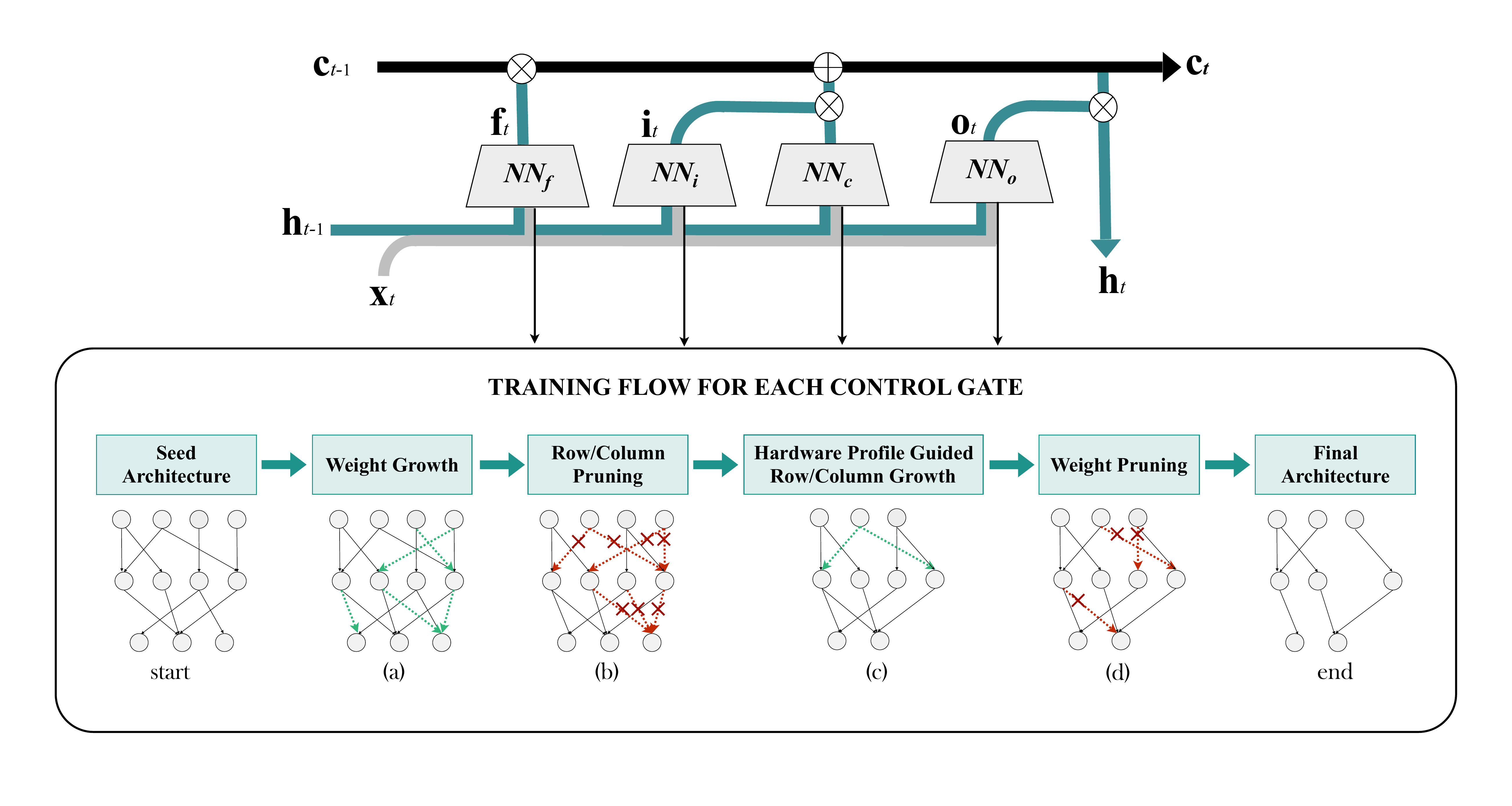}
\end{center}
\caption{Schematic diagram of our training methodology.}
\label{fig:main_fig}
\end{figure*}

We illustrate the details of our training methodology in 
Fig.~\ref{fig:main_fig}. As shown in its upper section, we adopt the H-LSTM 
cell that adds hidden layers to its control gates~\cite{hlstm}. Its internal 
computation flow is governed by the following equations: 

\begin{equation*}
{\scriptsize
\begin{split}
 \left(
 \begin{matrix}
   \textbf{f}_{t}  \\
   \textbf{i}_{t}  \\
   \textbf{o}_{t}  \\
   \textbf{g}_{t}  
  \end{matrix}
  \right) 
  &=
   \left(
 \begin{matrix}
   NN_{f}([\textbf{x}_{t},\textbf{h}_{t-1}])  \\
   NN_{i}([\textbf{x}_{t},\textbf{h}_{t-1}])  \\
   NN_{o}([\textbf{x}_{t},\textbf{h}_{t-1}])  \\
   NN_{g}([\textbf{x}_{t},\textbf{h}_{t-1}])  
  \end{matrix}
  \right) 
  =
 \left(
 \begin{matrix}
   \sigma(\textbf{W}_f H^{*}( [\textbf{x}_{t},\textbf{h}_{t-1}])+\textbf{b}_f) \\
   \sigma(\textbf{W}_i H^{*}( [\textbf{x}_{t},\textbf{h}_{t-1}])+\textbf{b}_i) \\
   \sigma(\textbf{W}_o H^{*}( [\textbf{x}_{t},\textbf{h}_{t-1}])+\textbf{b}_o) \\
   tanh(\textbf{W}_g H^{*}( [\textbf{x}_{t},\textbf{h}_{t-1}])+\textbf{b}_g)
  \end{matrix}
  \right)  \\
  &\ \ \ \ \ \ \ \ \ \ \ \ \ \ \ \ \ \ \ \ \ \ \ \ \ \ \ \  \textbf{c}_{t} = \textbf{f}_{t} \otimes \textbf{c}_{t-1} + \textbf{i}_{t} \otimes \textbf{g}_{t}\\
  &\ \ \ \ \ \ \ \ \ \ \ \ \ \ \ \ \ \ \ \ \ \ \ \ \ \ \ \ \ \ \ \  \textbf{h}_{t} = \textbf{o}_{t} \otimes tanh(\textbf{c}_t)
\end{split}
}
\end{equation*}

\noindent
where $\textbf{f}_{t}$, $\textbf{i}_{t}$, $\textbf{o}_{t}$, $\textbf{g}_{t}$, 
$\textbf{x}_{t}$, $\textbf{h}_{t}$, and $\textbf{c}_{t}$ denote the forget 
gate, input gate, output gate, cell update vector, input, hidden state, and 
cell state at step $t$, respectively; $\textbf{h}_{t-1}$ and 
$\textbf{c}_{t-1}$ refer to the previous hidden and cell states at step 
$t-1$; $NN$, $H$, $\textbf{W}$, $\textbf{b}$, $\sigma$, and $\otimes$ refer 
to NN gates, hidden layer that performs a linear 
transformation followed by an activation function, weight matrix, bias, 
$sigmoid$ function, and element-wise multiplication, respectively; 
$^{*}$ indicates zero or more $H$ layers for each NN gate. $H$ layers offer 
three advantages. First, they enhance gate control through a multi-level 
abstraction, hence alleviate H-LSTM's reliance on external stacking. Second, 
they can be easily regularized through dropout, and thus lead to better 
generalization. Third, they offer a wide range of choices for internal 
activation functions, such as the ReLU. This may provide additional benefits, 
such as faster learning and computation reduction due to zero 
outputs~\cite{hlstm}.

We utilize four training steps to learn the values, connectivity, and 
dimensions of the NN gates in the H-LSTM. We show these steps in the lower part 
of Fig.~\ref{fig:main_fig}. Training starts from a sparse seed architecture 
that contains a small fraction of connections to facilitate the initial back 
propagation of gradient information. During the weight growth (wg) phase, it 
iteratively wakes up only the most effective connections to reach high accuracy 
based on the gradient information. Then, it uses structured row/column pruning 
(rcp) algorithms to shrink the network dimensions, leading to lower 
inference latency. Next, it profiles the latency model on hardware and uses
row/column growth (rcg) algorithms to obtain a network from LHE-aware locally 
optimal design points. This enables simultaneous latency and accuracy gains, 
as shown later. Finally, it prunes away some weights for extra compactness. 

We next explain in detail the algorithms involved in these four steps. Unless otherwise stated, we assume a mask-based approach for tackling sparse
networks. Each weight matrix \textbf{W} has a corresponding binary mask 
matrix \textbf{Msk} that is of the exact same size. We finally update 
each \textbf{W} with its corresponding \textbf{W}$\otimes$\textbf{Msk} after 
the training flow terminates.

\subsection{Weight growth \& pruning}

The main objective of the weight growth phase is to locate only the most 
effective dormant connections to reduce the value of the loss function $L$. 
We first evaluate $\partial L / \partial w$ for each dormant connection $w$ 
based on its average gradient over the entire training set. Then, we activate 
a dormant connection based on the following policy: 
\begin{equation*}
\scriptsize
\begin{array}{l}
    \textbf{Msk}(w) = 1,~\text{iff}~|w.grad| \geq (100(1-g_w))^{th}~
    \text{percentile of}~|\textbf{W}.grad| 
\end{array}
\end{equation*}
where $g_w$ denotes the weight growth ratio. This rule was first proposed 
in~\cite{nest}. It caters to dormant connections that are most efficient 
at loss function $L$ reduction, and enables the network to reach a target 
accuracy with far less redundancy than a fully connected model. This offers 
an accurate yet irredundant model for all the subsequent steps to act on.

We adopt the magnitude-based weight pruning strategy for final redundancy 
removal. Pruning of insignificant weights is an iterative process. In each 
iteration, we adopt the following policy for weight selection:
\begin{equation*}
\scriptsize
\begin{array}{l}
    \textbf{Msk}(w) = 0,~\text{iff}~|w| \leq (100p_w)^{th}~\text{percentile of}~|\textbf{W}| 
\end{array}
\end{equation*}
where $p_w$ denotes the weight pruning ratio. We prune a neuron if all its 
input (or output) connections are pruned away. We retrain the network after 
the weight pruning iteration to recover its performance. The pruning phase 
terminates when retraining cannot achieve a pre-defined accuracy threshold. 
In the final training step, weight pruning minimizes the memory requirement of 
the final inference model. It also provides a high weight sparsity level for
sparsity-driven libraries, such as Intel Math Kernel Library (MKL)~\cite{mkl} 
on Intel CPUs, as shown later.

\subsection{Row/column growth \& pruning}

The grow-and-prune approach at the row/column level enables the network to 
adaptively expand and shrink its dimensions. This leads to an effective 
descent in the model architecture space towards fast, accurate, yet 
execution-efficient design points. However, due to the introduction of a 
large number of sub-optimal design points by hardware, the model architecture 
space may become rather non-monotonic, thus necessitating a
carefully-crafted stopping criterion for this process.

\begin{algorithm}[t]
   \caption{Row/column pruning algorithm}
\label{algo:rc_pruning}
\begin{algorithmic}
\footnotesize
   \STATE {\bfseries Input:} $p_c$ - column pruning ratio, $p_r$ - row pruning ratio, $\textbf{W} \in R^{M\times N}$ - control gate tensor , $\textbf{Msk}\in R^{M\times N}$ - mask tensor
    \STATE {\bfseries Denote:} $M$ - row count, $N$ - column count
   \STATE $th_{r}$ = $(p_rN)^{th}$ smallest value in $sum(abs(\textbf{W}), axis = 1)$
   \STATE $th_{c}$ = $(p_cM)^{th}$ smallest value in $sum(abs(\textbf{W}), axis = 0)$

   \FOR {$1\leq m\leq M$}
   \IF {$sum(abs(\textbf{W}_{m,:})) \leq th_{c}$}
   \STATE $\textbf{Msk}_{m,:} = 0$
   \STATE $\textbf{W}_{m,:} = 0$
   \ENDIF
   \ENDFOR

   \FOR {$1\leq n\leq N$}
   \IF {$sum(abs(\textbf{W}_{:, n})) \leq th_{r}$}
   \STATE $\textbf{Msk}_{:, n} = 0$
   \STATE $\textbf{W}_{:, n} = 0$
   \ENDIF
   \ENDFOR
\end{algorithmic}
\end{algorithm}

\begin{algorithm}[t]
   \caption{Row/column growth algorithm}
\label{algo:rc_growth}
\begin{algorithmic}
\footnotesize
   \STATE {\bfseries Input:} $g_c$ - column growth ratio, $g_r$ - row growth ratio, $\textbf{W} \in R^{M\times N}$ - control gate tensor , $\textbf{Msk}\in R^{M\times N}$ - mask tensor  \\
   \STATE {\bfseries Denote:} $M$ - row count, $N$ - column count, $\textbf{G} 
   \in R^{M\times N}$ - bridging gradient matrix, $set_c$ - indices for existing columns, $set_r$ - indices for existing rows, $lr$ - current learning rate

   \FOR {$1\leq m\leq M, 1\leq n\leq N$}
   \IF {${m} ~\text{in}~ set_c ~\text{\textbf{and}}~ {n} ~\text{in}~ set_r$}
   \STATE $\textbf{G}_{m,n} = 0$ 
   \ENDIF
   \ENDFOR

   \STATE $th_{r}$ = $(g_rN)^{th}$ largest value in $sum(abs(\textbf{G}_{:,set_c}), axis = 1)$   
   \STATE $th_{c}$ = $(g_cM)^{th}$ largest value in $sum(abs(\textbf{G}_{set_r,:}), axis = 0)$  

   \FOR {$1\leq m\leq M$}
   \IF {$sum(abs(\textbf{G}_{m,set_c})) \geq th_{r}$}
   \STATE $\textbf{Msk}_{m,set_c} = 1$
   \STATE $\textbf{W}_{m, set_c} = \textbf{G}_{m,set_c}\cdot lr$
   \ENDIF
   \ENDFOR

   \FOR {$1\leq n\leq N$} 
   \IF {$sum(abs(\textbf{G}_{set_r, n})) \geq th_{c}$}
   \STATE $\textbf{Msk}_{set_r, n} = 1$
   \STATE $\textbf{W}_{set_r, n} = \textbf{G}_{set_r,n}\cdot lr$
   \ENDIF
   \ENDFOR
\end{algorithmic}
\end{algorithm}

We propose row/column grow-and-prune algorithms to exploit LHE for 
hardware-symbiotic solutions.  The pruning algorithm takes advantage of the 
global trend to shrink the model dimension for latency reduction, whereas 
the growth algorithm recovers the model back to its corresponding LHP for 
simultaneous latency and accuracy gains. 

We present the row/column pruning algorithm in Algorithm~\ref{algo:rc_pruning}. 
Inspired by magnitude-based weight pruning methods, we examine the sum of 
the magnitudes of the weights per row/column for importance ranking. 
Row/column pruning is also an iterative process. We retrain the network 
after each pruning iteration, and stop if retraining cannot recover 
the performance to a pre-defined accuracy threshold. 

Algorithm~\ref{algo:rc_growth} illustrates our gradient-based row/column growth 
algorithm. Similar to the weight growth algorithm, we first evaluate 
$\partial L / \partial w$ for all the dormant connections in the network 
based on the average gradient over the entire training set (or a large 
batch). We only wake up the dormant rows and columns that possess the largest 
gradient magnitude sums, hence yielding the most efficiency in the reduction of 
loss function $L$. 

\begin{table*}[h]
\centering
\caption{Performance gain breakdown of each training step for language modeling on GPUs}
\begin{tabular}{clccccc}
\hline
\multirow{2}{*}{Step} & \multirow{2}{*}{Model}   & \multirow{2}{*}{\#Params.} & \multirow{2}{*}{Perplexity} & \multicolumn{3}{c}{Latency}\\
       &  &         &             & P2000 & P100 & V100   \\
\hline
& Baseline LSTM       & 14.46M & 72.1       & 3.52ms  & 2.12ms & 2.72ms \\
(a) & H-LSTM+wg            & \;\;4.68M  & 70.2   & 2.21ms  & 1.45ms & 1.81ms \\
(b) & H-LSTM+wg+rcp        & \;\;3.21M  & 72.2   & 1.78ms  & 1.27ms & 1.51ms \\
(c) & H-LSTM+wg+rcp+rcg    & \;\;3.24M  & 71.8   & 1.65ms  & 1.18ms & 1.14ms \\
(d) & H-LSTM+wg+rcp+rcg+wp & \;\;0.80M & 72.1    & 1.65ms  & 1.18ms & 1.14ms \\
\hline
\label{tb:breakdown}
\end{tabular}
\end{table*}

\begin{figure*}[h]
\begin{center}
\includegraphics[width = \linewidth]{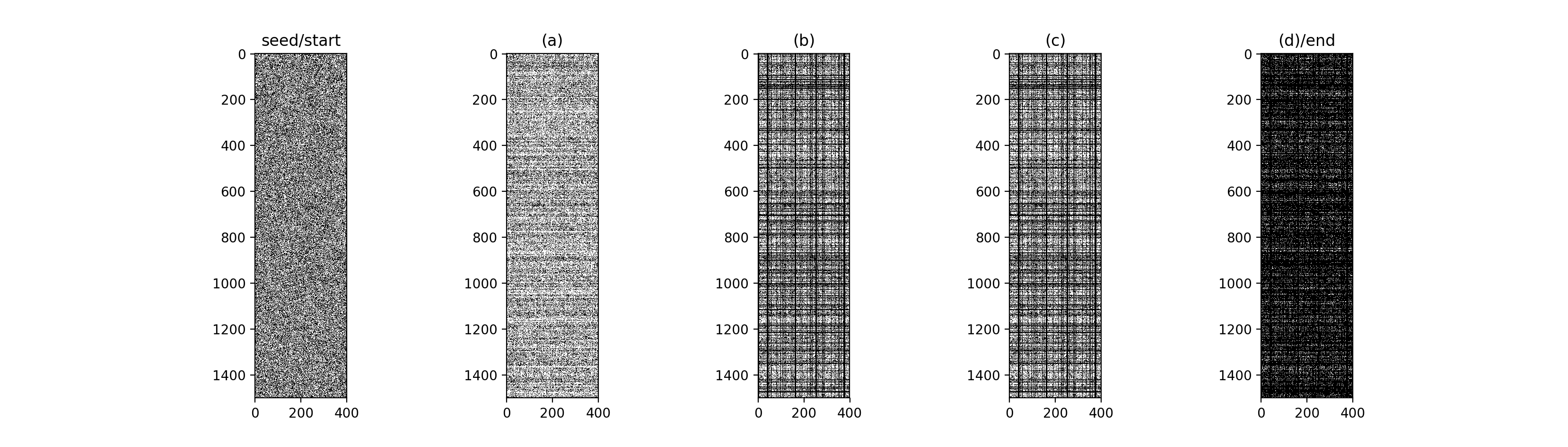}
\end{center}
\caption{Change of a typical input-to-hidden-layer weight matrix in the 
H-LSTM in different phases.  White sections denote active connections. All 
non-zero elements are changed to 1 for better visualization.}
\label{fig:phases}
\end{figure*}

\section{Experimental Results}

We next present our experimental results for the language modeling and speech 
recognition benchmarks. We implement our methodology using 
PyTorch~\cite{pytorch} on both Nvidia GPUs and Intel Xeon CPUs. For GPU, 
we have experimented with Nvidia GPUs: Quadro P2000, Tesla P100, and Tesla 
V100. For CPU, we have targeted Intel Xeon CPUs: Gold 5118 (2.3 GHz), 
E5-2682 v4 (2.5 GHz), and Broadwell (2.4 GHz). For CPU inference, we use 
Intel MKL~\cite{mkl} implementations for sparse matrix operation. 

\subsection{Language modeling}

We first demonstrate the effectiveness of our approach on language modeling. 

\textbf{Model architecture:} We experiment with a stacked LSTM architecture 
for this application that feeds embedded word vectors to the recurrent layers. 
The word vocabulary has size 10,000.  The dimension of the input word 
embedding is 400. We first train a conventional stacked LSTM architecture as 
the baseline. It contains two stacked recurrent layers, each with the 
hidden state width set to 1500, same as in~\cite{wenwei,zhu,stanford}. Next, 
we implement our methodology on a one-layer H-LSTM with the hidden state 
width again set to 1500. Each control gate contains one hidden layer with 
this width. 

\textbf{Dataset:} We report results on the Penn Treebank (PTB) 
dataset~\cite{ptb}. It contains 929k, 73k, and 82k words in the training, 
validation, and test sets, respectively.

\textbf{Training:} We use a stochastic gradient descent (SGD) optimizer for this application. 
We initialize the learning rate to 30, decayed by 10 when the validation accuracy does not 
increase in 50 consecutive epochs. We use a batch size of 32 for training. 
We use a dropout ratio of 0.2 for the hidden layers in the 
control gates, as in~\cite{hlstm}, 0.65 for input embedding layers, and 0.1 
for input words, as in~\cite{sales}. We employ L2 regularization during 
training with a weight decay of $1.2\times10^{-6}$. We use word-level 
perplexity as our evaluation criterion, same as in \cite{wenwei,zhu,stanford}.

We next present our experimental results for GPU and CPU inference.

\begin{table*}[t]
\centering
\caption{Inference model comparison for language modeling on GPUs}
\begin{tabular}{llcccc}
\hline
 \multirow{2}{*}{Model}   & \multirow{2}{*}{\#Params.} & \multirow{2}{*}{Perplexity} & \multicolumn{3}{c}{Latency}\\
        &         &             & P2000 & P100 & V100   \\
\hline
Our stacked LSTM - baseline & 14.5M    &  72.1       &  3.52ms & 2.12ms & 2.72ms \\
Sukhbaatar et al.~\cite{memn2n}       & -         & 120       & - & - & -\\
Mikolov et al.~\cite{mikolov}      & -         & 115/115   & - & - & -\\
Wen et al.~\cite{wenwei}       & 14.9M     & 78.7      & - & - & -\\
Zhu and Gupta~\cite{zhu}          & \; 7.2M      & 77.5      & 7.94ms*  & 3.69ms* & 3.02ms* \\
Zaremba, Sutskever, and Vinyals~\cite{zaremba}      & 18.0M$^+$    & 73.6      & - & - & -\\
Lin et al.~\cite{stanford}     & \; 9.0M$^+$     & 72.2      & - & - & -\\
\hline
\textbf{This work}     & \textbf{0.80M}~\textbf{(18.0$\times$)} & \textbf{72.1} & \textbf{1.65ms}~\textbf{(2.1$\times$)}  & \textbf{1.18ms}~\textbf{(1.8$\times$)} & \textbf{1.14ms}~\textbf{(2.4$\times$)} \\
\hline
\multicolumn{6}{l}{\scriptsize{$^+$Pessimistic estimate at zero word embedding dimension, i.e.
ultimate lower bound, and 1500 hidden state width as reported in the paper.}}\\
\multicolumn{6}{l}{\scriptsize{$*$ Measured based on our implementation of the exact same configuration as reported in the paper. }}\\
\label{tb:lm_gpu}
\end{tabular}
\end{table*}

\subsubsection{Synthesized models for GPU inference}

We first implement our methodology on various GPUs and compare our results with those for the 
conventional stacked LSTM architecture in Table~\ref{tb:breakdown}. Latency indicates the average 
value per test word sequence of length 70 over the entire test set at a batch size of 16. It can 
be observed that the four steps in our learning algorithm work sequentially and collaboratively to 
learn structured sparsity in the network, as shown in Fig.~\ref{fig:phases}. To further illustrate 
the performance gains obtained in each training step, we also break down their individual 
contributions in Table~\ref{tb:breakdown}, and present the details next: 

Step (a): The seed architecture has a 50\% sparsity level. In the weight growth phase, we use a 
growth ratio set to 10\% for the first eight epochs. This enables the network to reach a 
70.2 perplexity with only 65\% of its available connections, i.e., at a 35\% sparsity level. 

Step (b): We use equal pruning ratios for rows and columns of 20\%. We halve the pruning ratios if 
the post-retraining perplexity surpasses a pre-defined performance threshold. For this application, 
we set the performance threshold to 72.1. This is the performance achieved by our stacked LSTM model, 
and better than the values reported in~\cite{zaremba,wenwei,stanford,zhu}. In the final stage, we 
iteratively prune away single rows and columns until the performance threshold can no longer be 
satisfied. This enables us to fully exploit the global monotonic trend for latency reduction. After 
this step, the dimension of each control gate matrix shrinks from 1500/400 to 1197/317. This brings 
a 14\% to 20\% reduction in inference latency. 

Step (c): We next locate the LHPs for each GPU before starting the growth process. For this 
application, we find all three GPUs favor the same 1200/320 LHP in the model architecture space
defined by the control gate dimension. We then calculate the growth ratios accordingly to recover 
the network into this LHP. As expected, LHE exploration enables a 7\% to 23\% reduction in measured 
inference latency jointly with a 0.4 perplexity bonus. 

Step (d): We use an initial weight pruning ratio of 70\% and update it based on the same rule as 
in Step (b).  This step further reduces the number of network parameters by 4.1$\times$. 

We compare our final inference model with relevant work in Table~\ref{tb:lm_gpu}. Our models 
outperform the ones in the literature from all three design perspectives. Against the stacked 
LSTM baseline, we reduce the number of parameters by 18.0$\times$, and measured run-time latency by 
2.1$\times$, 1.8$\times$, and 2.4$\times$ on Nvidia P2000, P100, and V100
GPUs, respectively, without any accuracy degradation.

\subsubsection{Synthesized models for CPU inference}

We next report results from the implementation of our methodology on CPUs. We base our experiments 
on the Intel MKL~\cite{mkl} implementation due to its support and acceleration of sparse matrix 
computations. For CPU inference, we skip the dimension reduction process
(i.e., Step(b) and 
Step(c)) to fully exploit the potential of weight sparsity. The high sparsity level without 
dimension reduction at 93.4\% enables us to fully explore the benefit of sparsity acceleration, 
as opposed to a sparsity level at 83.5\% with dimension reduction that undermines the benefits of 
MKL. This yields additional 1.6$\times$ latency and 2$\times$ parameter reduction. Our final model 
has a test perplexity of 72.1, same as the test perplexity of the LSTM baseline. It only contains 
0.47M parameters as opposed to the baseline LSTM model that has 14.5M parameters (leading to a 
30.5$\times$ compression ratio). 

\begin{table*}[h]
\centering
\caption{Inference model comparison for language modeling on CPUs}
\begin{tabular}{lccc}
\hline
CPU platform & Baseline LSTM latency & This work & Speed-up factor \\
\hline
Intel Xeon E5-2682 v4 & 115.90ms & 22.64ms & $5.1\times$ \\
Intel Xeon Gold 5118 & 125.08ms & 24.02ms & $5.2\times$ \\
Intel Xeon Broadwell & 81.03ms & 19.57ms & $4.1\times$ \\
\hline
\label{tb:lm_cpu}
\end{tabular}
\end{table*}

We next compare the latency of the final inference models on CPUs in Table~\ref{tb:lm_cpu}. Relative 
to the LSTM baseline for language modeling, we reduce the inference latency by 76.1\% (4.2$\times$), 
80.8\% (5.2$\times$), and 75.8\% (4.1$\times$) on Intel Xeon Gold 5518, E5-2682 v4, and
Broadwell CPUs, respectively. Sparsity-driven MKL acceleration contributes approximately 
2.5$\times$ speed-up, while utilizing H-LSTM cells contributes the remaining 2$\times$ speed-up on 
the CPUs.

\subsection{Speech recognition}

We now consider another well-known application: speech recognition.

\textbf{Model architecture:} We implement a bidirectional DeepSpeech2 architecture that employs 
stacked recurrent layers following the convolutional layers for speech 
recognition~\cite{deepspeech2}. We extract Mel-frequency cepstral coefficients from the speech data 
in a 20ms feature extraction window. There are two CNN layers present prior to the recurrent layers 
and one connectionist temporal classification layer for decoding~\cite{ctc} after the recurrent 
layers. The width of the hidden state is 800, same as in~\cite{stanford,github}. Each control gate 
contains one hidden layer with width 800. 

\textbf{Dataset:} We obtain the results for the AN4 dataset~\cite{an4}. It contains 948 training 
utterances and 130 testing utterances. 

\textbf{Training:} We utilize a Nesterov SGD optimizer in our experiment. We use a batch size of 
16 for training. We initialize the learning rate to $3$$\times$$ 10^{-4}$ and decay it by 0.99 after 
every training epoch. We use a dropout ratio of 0.2 for the hidden layers in the H-LSTM. We 
use batch normalization between recurrent layers. We use L2 regularization with a weight decay 
of $1\times10^{-4}$. We use word error rate (WER) as our evaluation criterion, same as 
in \cite{stanford,ethmit,github}. 

We adopt the model reported in \cite{stanford} as our LSTM baseline. It contains five stacked 
LSTM layers with a hidden state width of 800. Then, we implement our methodology 
and compare our results for GPU and CPU inference as follows.

\subsubsection{Synthesized models for GPU inference}
We summarize our results for GPU inference in Table~\ref{tb:breakdown_p2000}. Latency 
values indicate the average instance latency over the test set with a batch size of 16. 
In Table~\ref{tb:breakdown_p2000}, we also break down the changes in model characteristics 
throughout the training flow to separate the performance gains at each training step: 

\begin{table*}[h]
\centering
\caption{Performance gain breakdown of each training step for speech recognition on GPUs}
\begin{tabular}{cclccrrc}
\hline
GPU platform & Step & Model & \#Layers & Dimension & \#Params. & WER(\%) & Latency \\

\hline
     && Baseline LSTM~\cite{stanford}         & 5 & 800/800        & 50.4M & 12.90   & \;\;35.87ms* \\
\multirow{4}{*}{Nvidia P2000} & (a) & H-LSTM+wg              & 3 & 800/800       & 27.1M & 8.39    & 32.13ms \\
&(b) & H-LSTM+wg+rcp          & 3 & 626/626      & 18.2M & 10.29   & 22.55ms \\
&(c) & H-LSTM+wg+rcp+rcg      & 3 & 644/644      & 18.3M & 9.44    & 20.79ms \\
&(d) & H-LSTM+wg+rcp+rcg+wp   & 3 & 644/644     & 8.1M & 9.97   & 20.79ms \\
\hline
    && Baseline LSTM~\cite{stanford}         &  5 & 800/800 & 50.4M & 12.90 & \;\;24.04ms* \\
\multirow{4}{*}{Nvidia P100} &(a) & H-LSTM+wg              &  3 & 800/800 & 27.1M & 8.39 & 22.77ms \\
&(b) & H-LSTM+wg+rcp          &  3 & 626/626 & 18.2M & 10.30 & 19.61ms \\
&(c) & H-LSTM+wg+rcp+rcg      &  3 & 640/640 & 18.3M & 10.08 & 17.70ms \\
&(d) & H-LSTM+wg+rcp+rcg+wp   &  3 & 640/640 & 7.2M  & 10.25 & 17.70ms \\
\hline
    && Baseline LSTM~\cite{stanford}         &  5 & 800/800 & 50.4M & 12.90 & \;\;19.35ms* \\
\multirow{4}{*}{Nvidia V100} &(a) & H-LSTM+wg              &  3 & 800/800 & 27.1M & 8.39 & 17.67ms \\
&(b) & H-LSTM+wg+rcp          &  3 & 626/626 & 18.2M & 10.30 & 17.32ms \\
&(c) & H-LSTM+wg+rcp+rcg      &  3 & 640/640 & 18.3M & 10.08 & 13.99ms \\
&(d) & H-LSTM+wg+rcp+rcg+wp   &  3 & 640/640 & 7.2M  & 10.25 & 13.99ms \\
\hline
\multicolumn{7}{l}{\scriptsize{$*$ Measured based on our implementation of the exact same configuration as reported in the paper. }}
\label{tb:breakdown_p2000}
\end{tabular}
\end{table*}

\begin{table*}[h]
\centering
\caption{Inference model comparison for speech recognition on GPUs}
\begin{tabular}{l|c|c|c}
\hline
\multirow{2}{*}{Model} & \#Params. & WER(\%)  & Latency\\
 & P2000 / P100 / V100 & P2000 / P100 / V100 & P2000 / P100 / V100 \\
\hline
Lin et al.~\cite{stanford} - baseline & 50.4M & 12.90 & 35.87ms* / 24.04ms* / 19.35ms*\\
Alistarh et al.~\cite{ethmit}   & 13.0M & 18.85 & -\\
Naren~\cite{github}   & 37.8M & 10.52 & -\\
Dai, Yin, and Jha~\cite{hlstm}    & \; 2.6M  & 10.37 & 32.13ms / 22.77ms / 17.67ms \\
\hline
\textbf{This work}       & \textbf{8.1M} (\textbf{6.2}$\times$) / \textbf{7.2M} (\textbf{7.0}$\times$)  / \textbf{7.2M} (\textbf{7.0}$\times$) & \textbf{9.97} / \textbf{10.25} / \textbf{10.25} & \textbf{20.79ms} (\textbf{1.7}$\times$) / \textbf{17.70ms} (\textbf{1.4}$\times$) / \textbf{13.99ms} (\textbf{1.4}$\times$)\\
\hline

\label{tb:gpu_speech}
\end{tabular}
\end{table*}

Step (a): The seed architecture has a 50\% sparsity level. In the weight growth phase, we use a 
growth ratio of 10\% for the first six training epochs. This enables the network to reach an 
8.39\% WER with only 62\% of its available connections, i.e., at a 38\% sparsity level. 

Step (b): We also adopt equal row and column pruning ratios of 20\%, and update them using the same 
method as in Step (b) of the language modeling application. This trims down the dimension of each 
weight matrix from 800/800 to 626/626, thus reducing the measured inference latency by 
2\% to 30\% across the three targeted GPUs.

Step (c): Unlike in the case of language modeling, this step unveils different LHPs for the P2000 
and P100 GPUs: it recovers the network into a 640/640 LHP for the P2000 GPU and a 644/644 LHP for 
the P100 GPU.  LHE exploration for speech recognition enables an additional 0.22\% to 0.85\% WER 
reduction, jointly with a latency reduction of 9.2\% to 19.2\%. 

Step (d): We initialize the weight pruning ratio to 70\% and update it using the same rule as 
in Step (d) in the language modeling application. This step reduces the number of network parameters 
by 2.3$\times$ to 2.5$\times$. 

We compare our final inference models with relevant work in 
Table~\ref{tb:gpu_speech}. Our results outperform most of the previous work 
from all three design perspectives. Though containing more 
parameters than the models presented in~\cite{hlstm}, our models achieve 
higher accuracy and deliver substantial inference speed-ups. Relative to the 
conventional LSTM baseline~\cite{stanford}, our method reduces the measured 
inference latency by 1.7$\times$ (1.4$\times$/1.4$\times$) on the P2000 
(P100/V100) GPU, while simultaneously reducing the number of parameters by 
6.2$\times$ (7.0$\times$/7.0$\times$), and WER from 12.90\% to 9.97\% 
(10.25\%/10.25\%). 

\begin{table*}[h]
\centering
\caption{Inference model comparison for speech recognition on CPUs}
\begin{tabular}{lccccc}
\hline
CPU platform & Baseline LSTM latency & This work & Speed-up factor \\
\hline
Intel Xeon E5-2682 v4 & 179.0ms & 75.3ms & $2.4\times$ \\
Intel Xeon Gold 5118 & 202.1ms & 93.3ms & $2.2\times$ \\
Intel Xeon Broadwell  & 146.7ms & 66.3ms & $2.2\times$ \\
\hline
\multicolumn{4}{l}{\scriptsize{Baseline latency values are measured based on our implementation of the exact same configuration }}\\
\multicolumn{4}{l}{\scriptsize{as reported in~\cite{stanford}.}}
\label{tb:final_cpu_speech}
\end{tabular}
\end{table*}

\subsubsection{Synthesized models for CPU inference}

We also exploit weight sparsity for speech recognition on CPUs. Augmented by 
Intel MKL, weight sparsity offers substantial memory and latency reductions 
at run-time. Similar to language modeling, we skip the dimension reduction 
process (i.e., Step(b) and Step(c)) to fully exploit the 
potential of weight sparsity. The high sparsity level without dimension 
reduction at 94.2\% enables us to fully exploit sparsity acceleration, as 
opposed to a sparsity level at 88.8\% with dimension 
reduction that undermines the benefits of MKL. This yields additional 
1.4$\times$ latency and 2.8$\times$ parameter reduction. Our final CPU 
inference model contains only 2.6M parameters as opposed to the baseline 
LSTM model that has 50.4M parameters (19.4$\times$ compression ratio). It has 
a WER of 10.37$\%$, which is 2.53$\%$ more accurate than the LSTM baseline. 

We next compare the latency of the final inference models on the CPUs in 
Table~\ref{tb:final_cpu_speech}.  Relative to the LSTM baseline for speech 
recognition, we reduce the inference latency by 57.9\% (2.4$\times$), 
53.8\% (2.2$\times$), and 54.8\% (2.2$\times$) on Intel Xeon Gold 
5518, E5-2682 v4, and Broadwell CPUs, respectively. Sparsity-driven MKL 
acceleration contributes approximately 2$\times$ speed-up, whereas utilizing 
H-LSTM cells contributes the remaining 1.1$\times$ speed-up.

\section{Conclusions}
In this work, we proposed a hardware-guided symbiotic training methodology 
for compact, accurate, yet execution-efficient inference models. By leveraging 
hardware-impacted LHE and multi-granular grow-and-prune algorithms, we were 
able to reduce LSTM latency while increasing its accuracy. We evaluated our 
algorithms on the language modeling and speech recognition applications. 
Relative to the traditional stacked LSTM architecture obtained for the Penn 
Treebank dataset, we reduced the number of parameters by 18.0$\times$, and 
measured run-time latency by 2.1$\times$, 1.8$\times$, and 2.4$\times$ on the 
Nvidia P2000, P100, and V100 GPUs, respectively, without any accuracy 
degradation. We reduced the number of parameters by 30.5$\times$, and measured 
run-time latency by 5.1$\times$, 5.2$\times$, and 4.1$\times$ on the Intel 
Xeon E5-2682 v4, Gold 5518, and Broadwell CPUs, respectively, without any
accuracy degradation. Relative to the DeepSpeech2 architecture obtained from 
the AN4 dataset, we reduced the number of parameters by 7.0$\times$, WER 
from 12.9\% to 9.9\%, and measured run-time latency by 1.7$\times$, 
1.4$\times$, and 1.4$\times$ on the Nvidia P2000, P100 and V100 GPUs, 
respectively. We also reduced the number of parameters by 19.4$\times$, 
WER from 12.9\% to 10.4\%, and measured run-time latency by 2.4$\times$, 
2.2$\times$, and 2.2$\times$ on the Intel Xeon E5-2682 v4, Gold 5518, and 
Broadwell CPUs, respectively. Thus, our method yields compact, accurate, yet 
execution-efficient inference models.

\bibliographystyle{IEEEtran} 
\bibliography{bibib} 

\begin{thebibliography}{10}
\providecommand{\url}[1]{#1}
\csname url@samestyle\endcsname
\providecommand{\newblock}{\relax}
\providecommand{\bibinfo}[2]{#2}
\providecommand{\BIBentrySTDinterwordspacing}{\spaceskip=0pt\relax}
\providecommand{\BIBentryALTinterwordstretchfactor}{4}
\providecommand{\BIBentryALTinterwordspacing}{\spaceskip=\fontdimen2\font plus
\BIBentryALTinterwordstretchfactor\fontdimen3\font minus
  \fontdimen4\font\relax}
\providecommand{\BIBforeignlanguage}[2]{{%
\expandafter\ifx\csname l@#1\endcsname\relax
\typeout{** WARNING: IEEEtran.bst: No hyphenation pattern has been}%
\typeout{** loaded for the language `#1'. Using the pattern for}%
\typeout{** the default language instead.}%
\else
\language=\csname l@#1\endcsname
\fi
#2}}
\providecommand{\BIBdecl}{\relax}
\BIBdecl

\bibitem{deepspeech2}
D.~Amodei, R.~Anubhai, E.~Battenberg, C.~Case, J.~Casper, B.~Catanzaro,
  J.~Chen, M.~Chrzanowski, A.~Coates, G.~Diamos, E.~Elsen, J.~Engel, L.~Fan,
  C.~Fougner, T.~Han, A.~Hannun, B.~Jun, P.~LeGresley, L.~Lin, S.~Narang,
  A.~Ng, S.~Ozair, R.~Prenger, J.~Raiman, S.~Satheesh, D.~Seetapun,
  S.~Sengupta, Y.~Wang, Z.~Wang, C.~Wang, B.~Xiao, D.~Yogatama, J.~Zhan, and
  Z.~Zhu, ``{D}eep {S}peech 2 : {E}nd-to-{End} speech recognition in {English}
  and {Mandarin},'' in \emph{Proc. Int. Conf. Machine Learning}, vol.~48, 2016,
  pp. 173--182.

\bibitem{seq2seq}
I.~Sutskever, O.~Vinyals, and Q.~V. Le, ``Sequence to sequence learning with
  neural networks,'' in \emph{Proc. Advances in Neural Information Processing
  Systems}, 2014, pp. 3104--3112.

\bibitem{deepheart}
B.~Ballinger, J.~Hsieh, A.~Singh, N.~Sohoni, J.~Wang, G.~H. Tison, G.~M.
  Marcus, J.~M. Sanchez, C.~Maguire, J.~E. Olgin, and M.~J. Pletcher,
  ``{DeepHeart}: {S}emi-supervised sequence learning for cardiovascular risk
  prediction,'' in \emph{Proc. AAAI Conf. Artificial Intelligence.}, 2018, pp.
  2079--2086.

\bibitem{stanford}
Y.~Lin, S.~Han, H.~Mao, Y.~Wang, and W.~J. Dally, ``Deep gradient compression:
  {Reducing} the communication bandwidth for distributed training,''
  \emph{arXiv preprint arXiv:1712.01887}, 2017.

\bibitem{wenwei}
W.~Wen, Y.~He, S.~Rajbhandari, W.~Wang, F.~Liu, B.~Hu, Y.~Chen, and H.~Li,
  ``Learning intrinsic sparse structures within long short-term memory,''
  \emph{arXiv preprint arXiv:1709.05027}, 2017.

\bibitem{lstm}
S.~Hochreiter and J.~Schmidhuber, ``Long short-term memory,'' \emph{Neural
  Computation}, vol.~9, no.~8, pp. 1735--1780, 1997.

\bibitem{deepspeech1}
A.~Hannun, C.~Case, J.~Casper, B.~Catanzaro, G.~Diamos, E.~Elsen, R.~Prenger,
  S.~Satheesh, S.~Sengupta, A.~Coates, and A.~Y. Ng, ``Deep {S}peech: {Scaling}
  up end-to-end speech recognition,'' \emph{arXiv preprint arXiv:1412.5567},
  2014.

\bibitem{him}
H.~Yin, Z.~Wang, and N.~K. Jha, ``A hierarchical inference model for
  {Internet-of-Things},'' \emph{IEEE Trans. Multi-Scale Computing Systems},
  vol.~4, pp. 260--271, 2018.

\bibitem{ternary}
C.~Zhu, S.~Han, H.~Mao, and W.~J. Dally, ``Trained ternary quantization,''
  \emph{arXiv preprint arXiv:1612.01064}, 2016.

\bibitem{shift}
B.~Wu, A.~Wan, X.~Yue, P.~Jin, S.~Zhao, N.~Golmant, A.~Gholaminejad,
  J.~Gonzalez, and K.~Keutzer, ``{S}hift: {A} zero flop, zero parameter
  alternative to spatial convolutions,'' in \emph{Proc. IEEE Conf. Computer
  Vision and Pattern Recognition}, 2018, pp. 9127--9135.

\bibitem{iot_energy}
A.~O. Akmandor, H.~Yin, and N.~K.~Jha, ``Smart, secure, yet energy-efficient,
  {Internet-of-Things} sensors,'' \emph{IEEE Trans. Multi-Scale Computing
  Systems}, 2018.

\bibitem{multicoset}
H.~Yin, B.~H. Gwee, Z.~Lin, A.~Kumar, S.~G. Razul, and C.~M.~S. See, ``Novel
  real-time system design for floating-point sub-{N}yquist multi-coset signal
  blind reconstruction,'' in \emph{Proc. IEEE Int. Symp. Circuits and Systems},
  May 2015, pp. 954--957.

\bibitem{nest}
X.~Dai, H.~Yin, and N.~K. Jha, ``{NeST}: {A} neural network synthesis tool
  based on a grow-and-prune paradigm,'' \emph{arXiv preprint arXiv:1711.02017},
  2017.

\bibitem{PruningHS}
S.~Han, J.~Pool, J.~Tran, and W.~Dally, ``Learning both weights and connections
  for efficient neural network,'' in \emph{Proc. Advances in Neural Information
  Processing Systems}, 2015, pp. 1135--1143.

\bibitem{admm}
T.~Zhang, K.~Zhang, S.~Ye, J.~Li, J.~Tang, W.~Wen, X.~Lin, M.~Fardad, and
  Y.~Wang, ``{ADAM-ADMM}: {A} unified, systematic framework of structured
  weight pruning for {DNNs},'' \emph{arXiv preprint arXiv:1807.11091}, 2018.

\bibitem{ese}
S.~Han, J.~Kang, H.~Mao, Y.~Hu, X.~Li, Y.~Li, D.~Xie, H.~Luo, S.~Yao, Y.~Wang,
  H.~Yang, and W.~J. Dally, ``{ESE}: {Efficient} speech recognition engine with
  sparse {LSTM} on {FPGA},'' in \emph{Proc. ACM/SIGDA Int. Symp.
  Field-Programmable Gate Arrays}, 2017, pp. 75--84.

\bibitem{baiduprune}
S.~Narang, E.~Elsen, G.~Diamos, and S.~Sengupta, ``Exploring sparsity in
  recurrent neural networks,'' \emph{arXiv preprint arXiv:1704.05119}, 2017.

\bibitem{hlstm}
X.~Dai, H.~Yin, and N.~K. Jha, ``Grow and prune compact, fast, and accurate
  {LSTM}s,'' \emph{arXiv preprint arXiv:1805.11797}, 2018.

\bibitem{scalpel}
J.~Yu, A.~Lukefahr, D.~Palframan, G.~Dasika, R.~Das, and S.~Mahlke, ``Scalpel:
  {C}ustomizing {DNN} pruning to the underlying hardware parallelism,''
  \emph{ACM SIGARCH Computer Architecture News}, vol.~45, no.~2, pp. 548--560,
  2017.

\bibitem{gru}
K.~Cho, B.~Van~Merri{\"e}nboer, D.~Bahdanau, and Y.~Bengio, ``On the properties
  of neural machine translation: Encoder-decoder approaches,'' \emph{arXiv
  preprint arXiv:1409.1259}, 2014.

\bibitem{qrnn}
J.~Bradbury, S.~Merity, C.~Xiong, and R.~Socher, ``Quasi-recurrent neural
  networks,'' \emph{arXiv preprint arXiv:1611.01576}, 2016.

\bibitem{deftnn}
P.~Hill, A.~Jain, M.~Hill, B.~Zamirai, C.-H. Hsu, M.~A. Laurenzano, S.~Mahlke,
  L.~Tang, and J.~Mars, ``{DeftNN}: {A}ddressing bottlenecks for {DNN}
  execution on {GPU}s via synapse vector elimination and near-compute data
  fission,'' in \emph{Proc. IEEE/ACM Int. Symp. Microarchitecture}, 2017, pp.
  786--799.

\bibitem{mitenergy}
T.-J. Yang, Y.-H. Chen, and V.~Sze, ``Designing energy-efficient convolutional
  neural networks using energy-aware pruning,'' \emph{arXiv preprint
  arXiv:1611.05128}, 2016.

\bibitem{chamnet}
X.~Dai, P.~Zhang, B.~Wu, H.~Yin, F.~Sun, Y.~Wang, M.~Dukhan, Y.~Hu, Y.~Wu,
  Y.~Jia, P.~Vajda, M.~Uyttendaele, and N.~K. Jha, ``{C}ham{N}et: {T}owards
  efficient network design through platform-aware model adaptation,''
  \emph{arXiv preprint arXiv:1812.08934}, 2018.

\bibitem{mobilenetv2}
M.~Sandler, A.~Howard, M.~Zhu, A.~Zhmoginov, and L.-C. Chen, ``Inverted
  residuals and linear bottlenecks: {Mobile} networks for classification,
  detection and segmentation,'' \emph{arXiv preprint arXiv:1801.04381}, 2018.

\bibitem{mnasnet}
M.~Tan, B.~Chen, R.~Pang, V.~Vasudevan, and Q.~V. Le, ``{MnasNet}:
  {Platform-aware} neural architecture search for mobile,'' \emph{arXiv
  preprint arXiv:1807.11626}, 2018.

\bibitem{shufflenetv2}
N.~Ma, X.~Zhang, H.-T. Zheng, and J.~Sun, ``{ShuffleNet V2}: {Practical}
  guidelines for efficient {CNN} architecture design,'' \emph{arXiv preprint
  arXiv:1807.11164}, 2018.

\bibitem{narangblock}
S.~Narang, E.~Undersander, and G.~Diamos, ``Block-sparse recurrent neural
  networks,'' \emph{arXiv preprint arXiv:1711.02782}, 2017.

\bibitem{cusparse}
``cu{SPARSE} library,'' \emph{NVIDIA Corporation, Santa Clara, California},
  2018.

\bibitem{cublas}
``cu{BLAS} library,'' \emph{NVIDIA Corporation, Santa Clara, California}, 2018.

\bibitem{porple}
G.~Chen, B.~Wu, D.~Li, and X.~Shen, ``{PORPLE}: {A}n extensible optimizer for
  portable data placement on {GPU},'' in \emph{Proc. IEEE Int. Symp.
  Microarchitecture}, 2014, pp. 88--100.

\bibitem{effisha}
G.~Chen, Y.~Zhao, X.~Shen, and H.~Zhou, ``{E}ffi{S}ha: {A} software framework
  for enabling effficient preemptive scheduling of {GPU},'' \emph{ACM SIGPLAN
  Notices}, vol.~52, no.~8, pp. 3--16, 2017.

\bibitem{mkl}
``{Intel} math kernel library,'' \emph{Intel Corporation, Santa Clara,
  California}, 2018.

\bibitem{pytorch}
A.~Paszke, S.~Gross, S.~Chintala, G.~Chanan, E.~Yang, Z.~DeVito, Z.~Lin,
  A.~Desmaison, L.~Antiga, and A.~Lerer, ``Automatic differentiation in
  {PyTorch},'' \emph{NIPS Workshop Autodiff}, 2017.

\bibitem{zhu}
M.~Zhu and S.~Gupta, ``To prune, or not to prune: {E}xploring the efficacy of
  pruning for model compression,'' \emph{arXiv preprint arXiv:1710.01878},
  2017.

\bibitem{ptb}
M.~Marcus, G.~Kim, M.~A. Marcinkiewicz, R.~MacIntyre, A.~Bies, M.~Ferguson,
  K.~Katz, and B.~Schasberger, ``The {P}enn {T}reebank: {A}nnotating predicate
  argument structure,'' in \emph{Proc. Workshop Human Language Technology},
  1994, pp. 114--119.

\bibitem{sales}
S.~Merity, N.~S. Keskar, and R.~Socher, ``An analysis of neural language
  modeling at multiple scales,'' \emph{arXiv preprint arXiv:1803.08240}, 2018.

\bibitem{memn2n}
S.~Sukhbaatar, A.~Szlam, J.~Weston, and R.~Fergus, ``End-to-end memory
  networks,'' in \emph{Proc. Advances in Neural Information Processing
  Systems}, 2015, pp. 2440--2448.

\bibitem{mikolov}
T.~Mikolov, A.~Deoras, S.~Kombrink, L.~Burget, and J.~{\v{C}}ernock{\`y},
  ``Empirical evaluation and combination of advanced language modeling
  techniques,'' in \emph{Proc. Annual Conf. Int. Speech Communication
  Association}, 2011, pp. 605--608.

\bibitem{zaremba}
W.~Zaremba, I.~Sutskever, and O.~Vinyals, ``Recurrent neural network
  regularization,'' \emph{arXiv preprint arXiv:1409.2329}, 2014.

\bibitem{ctc}
A.~Graves, S.~Fern{\'a}ndez, F.~Gomez, and J.~Schmidhuber, ``Connectionist
  temporal classification: {L}abelling unsegmented sequence data with recurrent
  neural networks,'' in \emph{Proc. Int. Conf. Machine Learning}, 2006, pp.
  369--376.

\bibitem{github}
S.~Naren, ``Speech recognition using {DeepSpeech}2,''
  \url{https://github.com/SeanNaren/deepspeech.pytorch/releases}, 2018.

\bibitem{an4}
A.~Acero, ``Acoustical and environmental robustness in automatic speech
  recognition,'' in \emph{Proc. IEEE Int. Conf. Acoustics, Speech, and Signal
  Processing}, 1990.

\bibitem{ethmit}
D.~Alistarh, D.~Grubic, J.~Li, R.~Tomioka, and M.~Vojnovic, ``{QSGD}:
  {C}ommunication-efficient {SGD} via gradient quantization and encoding,'' in
  \emph{Proc. Advances in Neural Information Processing Systems}, 2017, pp.
  1709--1720.

\end{thebibliography}

\end{document}